\documentclass[letterpaper, 10 pt, conference]{ieeeconf}
\IEEEoverridecommandlockouts                              % This command is only needed if 
\usepackage{cite}
\usepackage{multirow}
\usepackage{algorithm}  	
\usepackage{algorithmic}
\usepackage{graphicx}
\usepackage{times}
\usepackage{amsmath}
\usepackage{amssymb}
\usepackage{subfigure}
\usepackage{graphics} % for pdf, bitmapped graphics files
\usepackage[utf8]{inputenc}
\usepackage{boldline}
\usepackage{multicol}
\usepackage{bbm}

\usepackage[pdfstartview=FitH,bookmarksnumbered=true,colorlinks,bookmarksopen=true]{hyperref}
\newcommand{\eg}{\textit{e}.\textit{g}.~}
\title{\LARGE \bf
Fault-Aware Robust Control via Adversarial Reinforcement Learning}

\author{Fan Yang, Chao Yang, Di Guo, Huaping Liu, Fuchun Sun
\thanks{The authors are with the Department of Computer Science and Technology,
Tsinghua University, Beijing 100084, China, and also with the State
Key Laboratory of Intelligent Technology and Systems, Beijing National
Research Center for Information Science and Technology, Tsinghua University,
Beijing 100084, China (e-mail: hpliu@tsinghua.edu.cn).}}
\begin{document}

\maketitle
\thispagestyle{empty}
\pagestyle{empty}

%%%%%%%%%%%%%%%%%%%%%%%%%%%%%%%%%%%%%%%%%%%%%%%%%%%%%%%%%%%%%%%%%%%%%%%%%%%%%%%%
\begin{abstract}
 Robots have limited adaptation ability compared to humans and animals in the case of damage. However, robot damages are prevalent in real-world applications, especially for robots deployed in extreme environments. The fragility of robots greatly limits their widespread application. We propose an adversarial reinforcement learning framework, which significantly increases robot robustness over joint damage cases in both manipulation tasks and locomotion tasks. The agent is trained iteratively under the joint damage cases where it has poor performance. We validate our algorithm on a three-fingered robot hand and a quadruped robot. Our algorithm can be trained only in simulation and directly deployed on a real robot without any fine-tuning. It also demonstrates exceeding success rates over arbitrary joint damage cases. 
\end{abstract}
\section{Introduction}
%1. 从Joint Malfunction 角度出发介绍问题的重要性
Humans and animals can quickly adapt to limb and joint injuries\cite{cully2015robots}. For example, if the index finger is injured for a human, he would consciously avoid using the index finger and rely more on other fingers. We do not need to re-learn the methods to use our hands. Instead, we only need a few trials and would quickly adapt to a suitable way of manipulation. On the contrary, traditional robot algorithms can hardly adapt to any damage. Minor damage could lead to a breakdown of an entire robot system.

On the other hand, robots have demonstrated their efficacy in a wide range of areas in our society\cite{honarpardaz2017finger, chik2016review, khatib2016ocean, mahony2012multirotor}. However, currently, robots are mostly deployed in factories and research labs. One of the major obstacles for its widespread application in more diverse environments arises from its fragility, especially for complex robots\cite{carlson2005ugvs, atkeson2015no, chatzilygeroudis2018reset}. Not only robots in the laboratory, but even mature robot commercial products can also suffer greatly from malfunctions\cite{borden2007mechanical, kaushik2010malfunction}. Under such scenarios, solving the problem of joint malfunction, an important part of robot damage cases, is of great importance. 

Reasons for robot vulnerability are manifold. Not only because it lacks pre-programmed experience in damage cases, but also arises from the vulnerability of Deep Neural Network (DNN) to attack and perturbations.
Minimal perturbations to a DNN could possibly lead to unpredictable results \cite{goodfellow2014explaining}. Deep Reinforcement Learning (DRL) algorithms also suffer from such vulnerability\cite{huang2017adversarial,yeow2019spatiotemporally,pattanaik2017robust,tessler2019action}. Joint malfunction can be considered as a kind of attack, which could possibly imperil the stability of robot system and break it down. 
% Thus, a traditional DRL algorithm is very likely to fail on joint malfunction scenarios due to vulnerability and lack of experience.

Our work focuses on the problem of robot damage, especially joint malfunction. Some previous work mentions this problem\cite{nagabandi2018learning, cully2015robots, chatzilygeroudis2018reset}. However, they all choose locomotion tasks. No one focuses on manipulation tasks. Even in their locomotion tasks, they use extremely stable robot platforms. These robots have far more legs than needed. Besides, the centers of mass for these robots are usually extremely low, and robot legs are widespread outside. They are also not likely to fall over even if a certain leg is disabled. Manipulation tasks usually do not have much redundancy. Moreover, a wrong decision in manipulation tasks would possibly limit the action space. For example, one finger could be stuck by others in in-hand manipulation. Locomotion tasks would also become much challenging if disabling one leg would substantially impair its stability. We would expect our algorithm can adapt to a new policy rather than just a minor fine-tuning in previous work. We would expect the robot could recover from catastrophic damage rather than a perturbation-like breakage. To the author's best knowledge, we are also the first to propose the robot damage problem in manipulation tasks.
\begin{figure}
\centering
    \includegraphics[width=8.5cm]{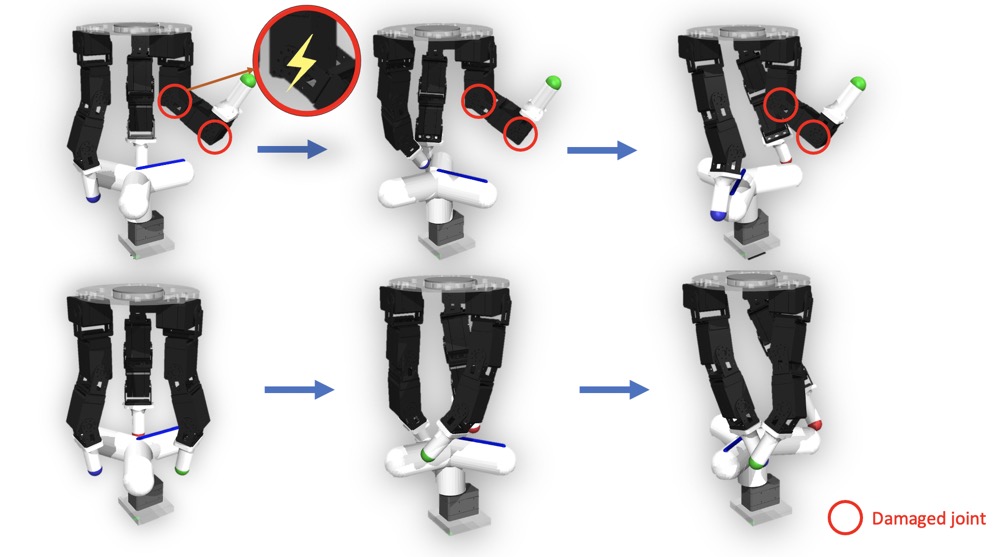}
    \caption{Robots are likely to suffer from damages in real-world applications. A minor damage may lead to the breakdown of the entire system. An algorithm that can enable the robot to adapt to different damage cases is of great importance. The three figures on the top show the robot with two damaged joints trying to turn the valve. The three figures on the bottom show the complete robot trying to turn the valve.}
    \label{fig:intro}
\end{figure}

Our contributions are three-fold:
\begin{enumerate}
    \item We propose and define the joint damage problem in manipulation tasks. 
    \item We develop an adversarial reinforcement learning framework for manipulation and locomotion tasks, which is able to adapt to different joint damage cases. Not only does our algorithm demonstrate exceeding robustness, it can also be trained only in simulation and directly deployed to real robots without any fine-tuning.
    % \item We develop an adversarial reinforcement learning framework, incorporate Soft Actor-Critic\cite{haarnoja2018soft} into our framework, and propose an algorithm which can solve joint damage cases. Our algorithm can be trained only in simulation directly deployed to real robots without any fine-tuning.
    \item We evaluate our algorithm on the D'Claw robot and the D'Kitty robot, validating the increased robustness over joint damages in simulation and real robots. Our algorithm also shows robustness over noise resistance.
\end{enumerate}
% \begin{enumerate}
%     \item We develop an adversarial reinforcement learning framework, which is capable of solving a great variety of joint damage cases in both manipulation tasks and locomotion tasks.
%     \item We incorporate Soft-Actor-Critic\cite{haarnoja2018soft} into our framework and propose an algorithm, which can solve joint damage cases. Our algorithm can be trained only in simulation directly deployed to real robots without any fine-tuning.
%     \item we evaluate our algorithm on D'Claw robot and D'Kitty robot, validating the increased robustness over joint damages in simulation and real robots. Our algorithm also demonstrates robustness over noise resistance.
% \end{enumerate}
\section{Related Works}

\label{sec:related_works}
% There are two possible perspectives of solving the robot damage problems: (1) increase the robustness of the control algorithm and (2) enable the policy to adapt from the source domain to the target damage domain. We refer to these two perspectives as robustness method and policy adaptation method, respectively.
%再看更多相关论文？
% \subsection{Robustness Method}
% \label{sec:robustness_method}
In \cite{peng2018sim, tobin2017domain, akkaya2019solving}, the authors propose domain randomization methods. Reinforcement Learning(RL) agent is iteratively trained in different randomized environments. Joint damage cases can be considered as an instance of the training environment set. In this method, the RL agent is agnostic to the change of environment parameters. This results in the fact that the RL agent will ultimately learn an averaged policy, which is likely to deal with any possible environment but probably has limited performances. More importantly, domain randomization methods do not guarantee the convergence of the RL training process. In some extreme cases, \eg randomizing joint working states, the policy will not converge or will only converge to an unsatisfying point if environments vary too much. 
% For instance, randomizing joint working states is likely to result in the convergence problem for it is a different and difficult kind of variation compared to friction, noise and mass, which usually randomize in traditional domain randomization algorithm. 

Robust Adversarial Reinforcement Learning (RARL) \cite{pinto2017robust} could be considered as a variant of domain randomization methods, in which the RL agent (protagonist) is trained iteratively in the worst environment. The worst environment is given by an adversary agent, which is trained along with the RL agent. RARL also suffers from the convergence problem mentioned for domain randomization. 
% Furthermore, it formulated the protagonist and the adversary as a zero-sum game. It would be hard to adjust the parameters to make sure the optimal equilibrium points guarantee a good performance.
RARL is the most similar to our work. We both use an adversarial training pipeline. However, our work differs greatly from RARL in the following aspects:
\begin{enumerate}
    \item The motivation of RARL is to increase robustness over unknown environment parameters and disturbances. While our work is trying to solve arbitrary joint damage cases by  sufficiently exploiting the shared information in each case.
    \item The convergence problem mentioned above is greatly relieved in our algorithm by giving joint working states to the RL agent.
    \item RARL and domain randomization methods only test robustness over ``soft constraints'', such as changed friction and mass. Our work also considers about ``hard constraints'', \eg joint damage, which restrict the action space of robots.
    \item RARL uses a parametric network to formalize the adversary while we use a parameter-free method. Details are discussed in \ref{sec:ad_learning_framework}.
\end{enumerate}

% \subsection{Policy Adaptation Method}
Another perspective of solving robot damage problems could be summarized as policy adaptation methods\cite{raileanu2020fast, huang2020one, nagabandi2018learning}. Policy adaptation methods usually incorporate Meta-learning methods in it. Joint working states are not implicitly available to the agent. The robot has to sample a few steps, estimate the changed dynamics, and adapt to the new environment based on its experience. 

However, these methods are extremely limited in our problem settings. The complexity of our tasks would make it extremely difficult to transfer the policy. Moreover, in order to gather sufficient experience to update and transfer policy, the agent has to be pre-trained under a number of different environments, which leads to computational complexity. \cite{raileanu2020fast} was tested in our problem, while it showed poor performances. 

% There are several limitations if we directly apply policy adaptation into the joint damage problem. Firstly, errors may accumulate in these consecutive processes, which limits the performance of the adapted policy. A self-diagnosis module would help get more accurate information about joint states, thus assisting a more efficient and accurate control policy. Secondly, In order to gather sufficient experience to update and transfer policy, the agent has to be pre-trained under a number of different environments, which leads to computational complexity. \cite{raileanu2020fast} was tested in our problem, while it showed poor performances. The complexity in policy and an insufficient number of training scenarios in our problem would make policy adaptation extremely difficult to learn.

% \subsection{Reinforcement Learning Algorithm}
On the other hand, the state-of-art algorithm Soft Actor-Critic (SAC) has exceeding performances in convergence speed and exploration process, which has the potential ability in helping solve robot damage problems.

SAC is an off-policy Actor-Critic RL algorithm based on the maximum entropy reinforcement learning framework. The Critic Network takes an action and a state as input and output a value evaluating the current state-action pair. While the Actor Network takes a state as input and is supposed to output the optimal action. The difference between a traditional Actor-Critic Network and SAC lies in its reward function. SAC jointly optimizes extrinsic reward from the environment and an entropy term to encourage a diverse exploration, resulting in a better exploration.

% More formally, its reward function is defined as the following equation:
% \begin{equation}
% J(\pi)=\sum_{t=0}^{T} \mathbb{E}_{\left(\mathbf{s}_{t}, \mathbf{a}_{t}\right) \sim \rho_{\pi}}\left[r\left(\mathbf{s}_{t}, \mathbf{a}_{t}\right)+\alpha \mathcal{H}\left(\pi\left(\cdot \mid \mathbf{s}_{t}\right)\right)\right] .
% \label{equ:sac}
% \end{equation}
% where $\mathcal{H}$ denotes the entropy term and $\alpha$ denotes the weight between the entropy and reward.

\section{Problem Formulation}
\label{sec:problem_formulation}

% \subsection{Reinforcement Learning Notations on Markov Decision Process}
% Reinforcement learning agents are designed to maximize the cumulative discounted reward in Markov Decision Processes(MDPs). 
% Concretely, in our paper, we examine MDPs represented by the following notations: $(\mathcal{S}, \mathcal{A}, \mathcal{P}, r, \gamma)$,
% where $\mathcal{S}$ is a set of states, $\mathcal{A}$ is a set of actions.
% $\mathcal{P}: \mathcal{S} \times \mathcal{A} \times \mathcal{S} \rightarrow \mathbb{R}$ is the transition probability. More specifically, it represents the probability of the next state $\mathbf{s}_{t+1} \in \mathcal{S}$ given the current state $\mathbf{s}_{t} \in \mathcal{S}$ and the action $\mathbf{a}_{t} \in \mathcal{A}$. 
% $r: \mathcal{S} \times \mathcal{A} \rightarrow \mathbb{R}$ means the reward function, and $\gamma$ is the discount factor.

% Our reinforcement algorithms are designed to learn a policy $\pi_{\theta}(\mathbf{a}_{t}|\mathbf{s}_{t})$, which is supposed to maximize the expectation of the sum discounted reward: $\sum_t \mathbb{E}(\gamma^{t}r(\mathbf{s}_t, \mathbf{a}_t)$ . The policy is defined by policy parameters: $\theta$

We focus on the problem of joint malfunction in robotic tasks. More specifically, joint malfunction in our task includes but is not limited to: (1) inability to change joint angles or (2) joints acting randomly. These are prevalent in real-world robotic tasks. For example, joints could get stuck or under-actuated so that further changing joint angles may not be possible; electronic errors could lead to joints acting randomly. In our problem settings, the number of joints that is possible to be damaged is not limited. As long as the task is still possible for the robot to complete, any number of joints damaged is acceptable. 

We define our robot system as a fault-aware system because we assume that joint working states are available to our robot in test scenarios. Many previous works focus on self-diagnosis modules \cite{kawabata2002study, zagal2009resilient, guan2015fault, yuan2011power}. Therefore, we mainly focus on the situation when the robot is aware of its damage situation.
% Our model consists of two parts: a self-diagnosis module and an RL policy module. Self-diagnosis modules detect whether the joint is functioning normally. Self-diagnosis modules for joints are prevalent, and self-diagnosis varies greatly in different robotic settings. For example, some motors may have a pre-programmed self-diagnosis module, while others may process joint angles or vision information. 
\begin{figure*}[htbp]
    \centering
    \includegraphics[width=16cm]{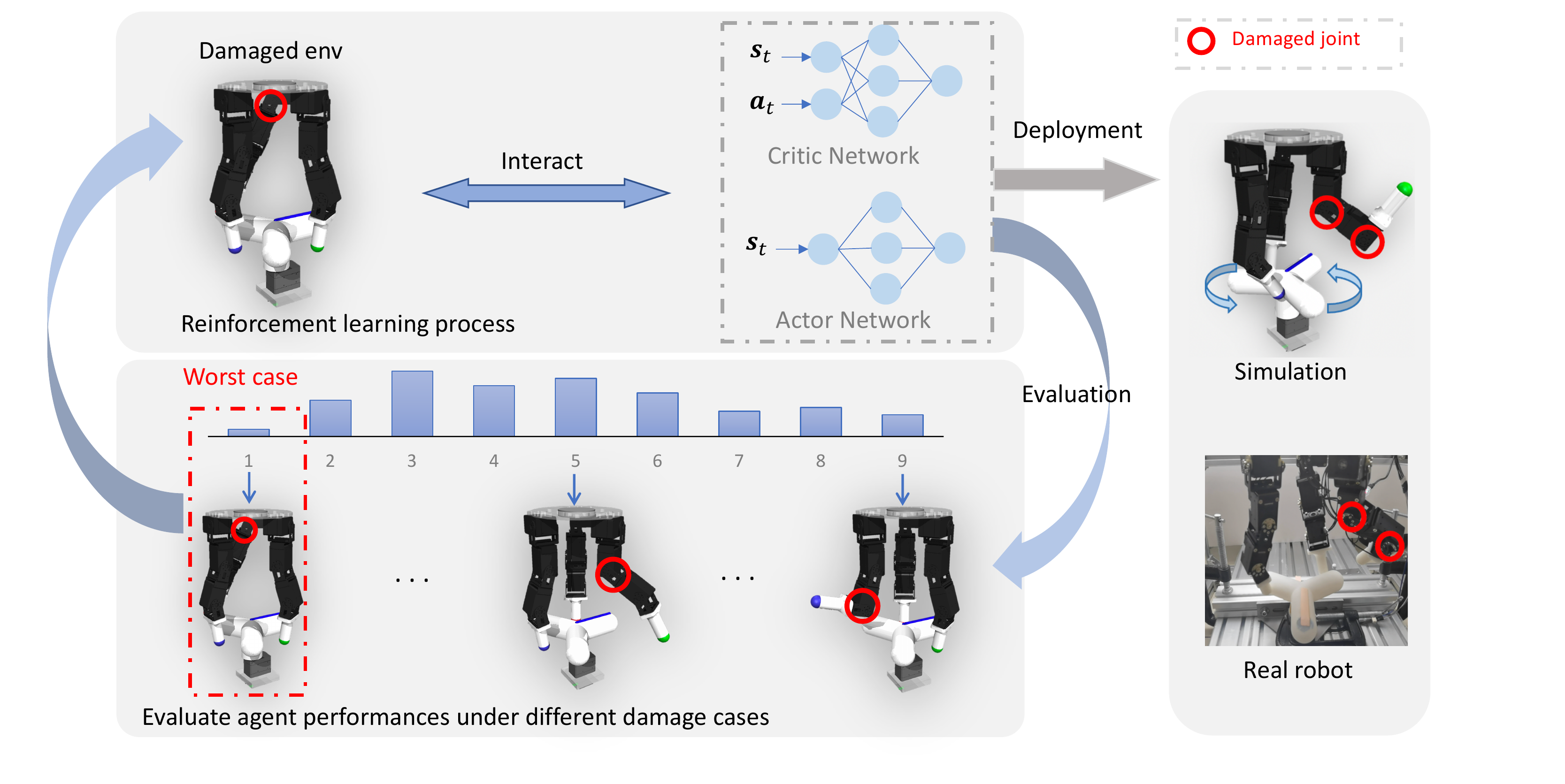}
    \caption{The framework of our method. We focus on the problem of joint damage in manipulation and locomotion tasks. In the training phase, there exists an adversarial process iteratively selecting the joint damage cases that can sabotage the agent performance most. The agent is always aware of the joint working states and iteratively trained under the environment the adversarial process returns. The agent trained in the method above demonstrates significantly increased robustness over joint damage and other perturbations. It also can be directly deployed to real-world applications.}
    \label{fig:abstract_fig}
\end{figure*}
More formally, we assume ground truth joint working states $\mathbf{q}$ are given. In our cases, $\mathbf{q}$ is a binary vector indicating whether each joint has been damaged. It could also be extended to other continuous variables such as temperature. We assume that robot action won't affect joint states. More formally, $\mathbf{q}$ is independent from $\mathbf{a}_{0:t}$. Note that we only need self-diagnosis in the testing processes. Because in training processes, damaged cases are defined by ourselves, the ground truth joint working states is always known. 

We formulate our control module as an RL algorithm, and the control process as a Markov Decision Process(MDP). For a traditional RL problem, the RL algorithm is designed to return a policy $\pi_{\theta}(\mathbf{a}_{t}|\mathbf{s}_{t})$ that maximizes the cumulative discounted reward $\sum_t \mathbb{E}(\gamma^{t}r(\mathbf{s}_t, \mathbf{a}_t))$  in MDP. But since we incorporate joint working states $\mathbf{q}$, we reform the policy as $\pi_{\theta}(\mathbf{a}_t|\mathbf{s}_t, \mathbf{q})$, using which the expected reward is maximized:
\begin{equation}
    \rho(\pi_{\theta};\theta) = \mathop{\mathbb{E}}\limits_{\mathbf{q}}\left[\mathbb{E}\left[\sum_{t} \gamma^tr(\mathbf{s}_t, \mathbf{a}_t)|\mathbf{s}_0, \theta, \mathbf{q} \right]\right],
    \label{equa:target}
\end{equation}
where $\mathbf{s}_t$ denotes the current state, $\mathbf{a}_t$ denotes the current action, $\theta$ denotes the parameter for the policy network, $\gamma$ is the discounted factor.

\section{Adversarial Learning Framework}
\label{sec:ad_learning_framework}
Standard deep RL algorithms usually suffer from limited robustness and vulnerability to perturbations. 
Specifically, a standard RL algorithm trained on a fully functional robot would possibly fail on those tasks with several joints disabled.

An intuitive algorithm to solve it would be training a distinct policy for each damage scenario.
However, due to various combinations of joints, the number of scenarios would increase exponentially as the number of joints increases, which makes it infeasible. We believe that there is shared information among those distinct policies. Though a traditional RL algorithm would fail on damage cases, the policy could potentially understand several basic information about those damage tasks, \eg the dynamic model, and the reward function. A minor fine-tuning would make it adapt to different damage cases. 
% The minor fine-tune may enable the policy to understand how the joint states would change the dynamics and how to plan under a new environment.

Moreover, as mentioned in Equation \ref{equa:target}, we are trying to optimize over all the joint working states, which is an extremely difficult task due to the large search space. Instead, we iteratively choose to optimize over scenarios where the agent has a poor performance. We call those scenarios challenging cases. It is probably because, in this case, the joints the current policy mostly relies on are damaged so that it becomes challenging. We believe that challenging cases may exposure the weakest point of the policy. Thus, because the policy for different joint states $\mathbf{q}$ is likely to share plenty of information, fixing challenging scenarios would not only improve the performance in it but also provide experiences for other cases. Furthermore, we believe increasing the performance under challenging cases will have a limited influence on other cases. Training iteratively on challenging cases would increase the average performance over all the cases because we are constantly optimizing the lower bound. More specifically, our algorithm works as follows: First, we train our policy on a predefined $\mathbf{q}$ for $K$ episodes. Next, we search for the challenging scenario under the updated policy and update $\mathbf{q}$ for the next policy training iteration. We iterative the processes above until converge. A pseudo-code is shown in Algorithm \ref{alg:rarl}.
\begin{algorithm}[htb]
\caption{Adversarial Learning Process}
\label{alg:rarl}
\begin{algorithmic}
\REQUIRE
    $N_{iter}$: The total number of epochs;
    $K$: The number of episodes to train the policy in each iteration;

\STATE Randomly initialize $\mathbf{q}, \theta$;
\STATE Damage the robot based on $\mathbf{q}$;
\STATE Initialize replay buffer $D$;

\FOR{$iter=1,2,...N_{iter}$}
    \FOR{$k=1,2,...K$}
        \STATE $\mathbf{\tau}' \longleftarrow \text{Roll}(\theta, \mathbf{q})$;
        \STATE $D \longleftarrow D \cup \{\mathbf{\tau}'\}$;
        \IF{Time to update}
        \STATE $\tau \longleftarrow \text{SampleTrajectories}(D)$;
        \STATE $\theta \longleftarrow \text{PolicyOptimizer}(\theta, \tau)$;
        \ENDIF
    \ENDFOR
    \STATE $\mathbf{q} \longleftarrow \text{SearchTheChallengingScenario}(\theta)$;
    \STATE Damage the robot based on $\mathbf{q}$
\ENDFOR
\RETURN $\theta$
\end{algorithmic}
\end{algorithm}

% TODO: is it wordy in this paragraph?
But how can we get the challenging $\mathbf{q}$? Ideally, it is supposed to be the worst case, in which the agent has the lowest reward. In order to get the worst case, searching over all the $\mathbf{q}$ would be inefficient. In RARL, the author proposed to use an adversary: a parametric network, to choose the worst $\mathbf{q}$. However, the action space for the adversary would be too large if we directly implement it. Moreover, for the adversary, it only needs to choose the worst damage cases, which is only a one-step process. The MDP is poorly formed for the adversary in this case. Instead, a parameter-free method of implementing the adversary is used in our algorithm, which suits joint damage problems much better. We apply a greedy search algorithm, as shown in Algorithm \ref{alg:greedy}.
\begin{algorithm}[htb]
\caption{Greedy Search the Challenging Scenario}
\label{alg:greedy}
\begin{algorithmic}
\REQUIRE
    $M$: The maximum number of joints that can be damaged;
    $N$: Total number of joints;
    $\theta$: Policy parameters;
    
\STATE Initialize the set of joints damaged in the challenging scenario: $U \longleftarrow \phi$
\FOR{$m=1,2,...M$}
    \FOR{$n=1,2,...N$}
    \STATE $\mathbf{q} \longleftarrow \text{DamageJoints}(U, n)$;
    \STATE$p_n \longleftarrow \text{Evaluate}(\theta, \mathbf{q})$;
    \ENDFOR
    \STATE $n_{min} \longleftarrow \min\limits_{n}\{p_n\}$
    % \STATE Choose the $\mathbf{q}_{n}$ with the minimal evaluation score;
    \STATE $U = U \cup \{n_{min}\}$;
\ENDFOR
\RETURN $U$
\end{algorithmic}
\end{algorithm}
Though a greedy search algorithm does not guarantee a worst scenario, we only need one that is challenging enough. Searching over the entire space is much more computationally expensive than greedy search, while it does not yield a substantially improvement in the performance.

Though in RARL, an adversarial training process is also implemented, the adversary action is never given to the agent. While in our method, the agent is always aware of its joint working states. Thus our agent is able to plan distinctive actions for different $\mathbf{q}$. This is important because it could avoid the problem mentioned in \ref{sec:related_works}. In fact, in our experiments, if $\mathbf{q}$ is not provided, the RL training process will not converge to a successful policy.

\section{Experiments}

\begin{figure*}[htbp]
    \centering
    \includegraphics[width=15cm]{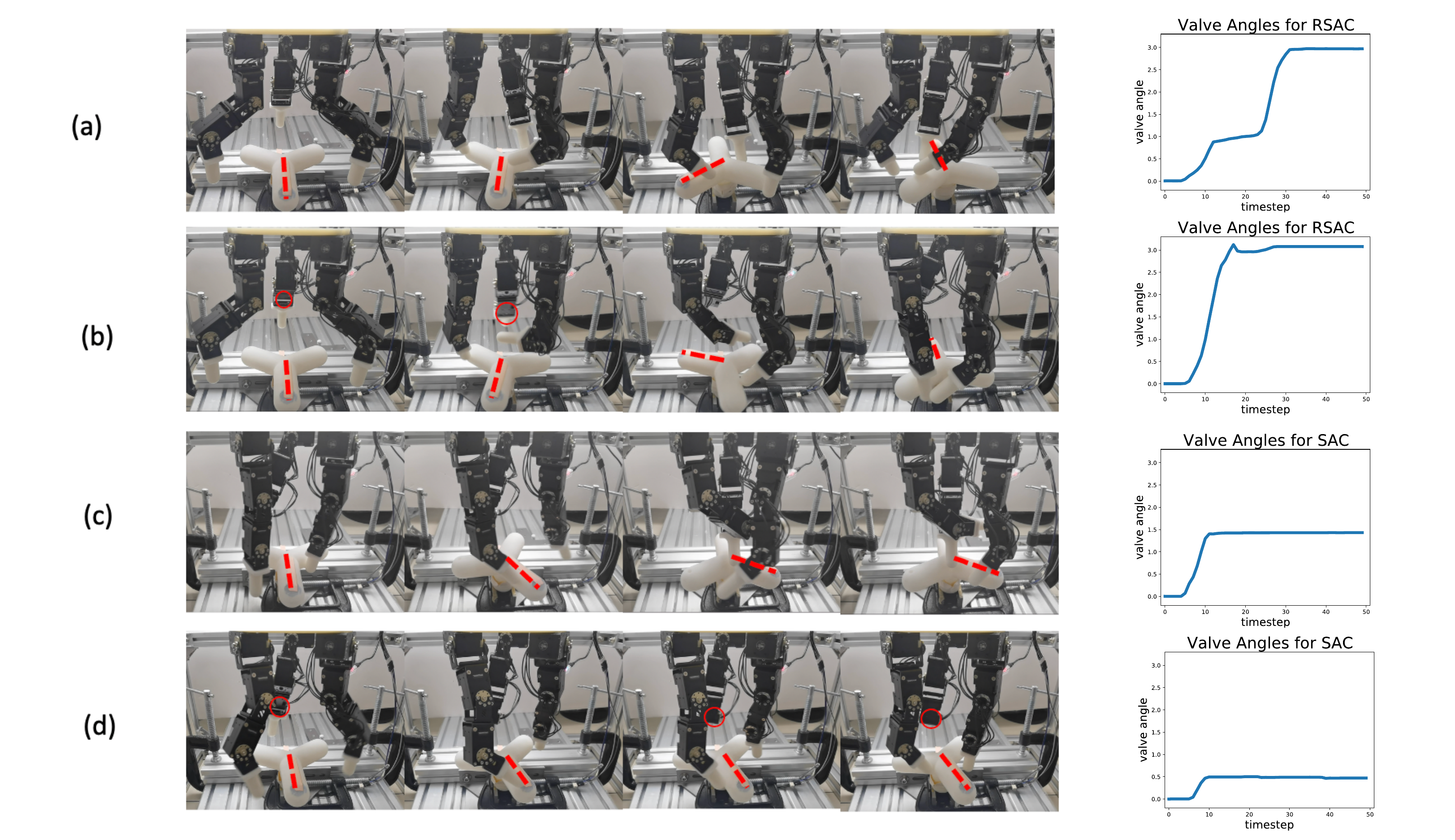}
    \caption{This figure shows photo samples of D'Claw manipulation experiments of single-joint-damaged cases and complete cases of it. Photos are not sampled with the same time intervals, but sampled for clarity. The valve angles in these cases are shown on the right side. The red circle in the figures denote the broken joint. The red dashed lines on the valves are used to mark the rotation. Figure (a) shows the RSAC method in a complete case. Figure (b) shows the RSAC method when a single joint is damaged. Figure (c) shows the SAC method in a complete case. Figure (d) shows the SAC method when a single joint is damaged. }
    \label{fig:big_exp}
\end{figure*}
\begin{figure}[htbp]
    \centering
    \subfigure[]{
        \includegraphics[width=4cm]{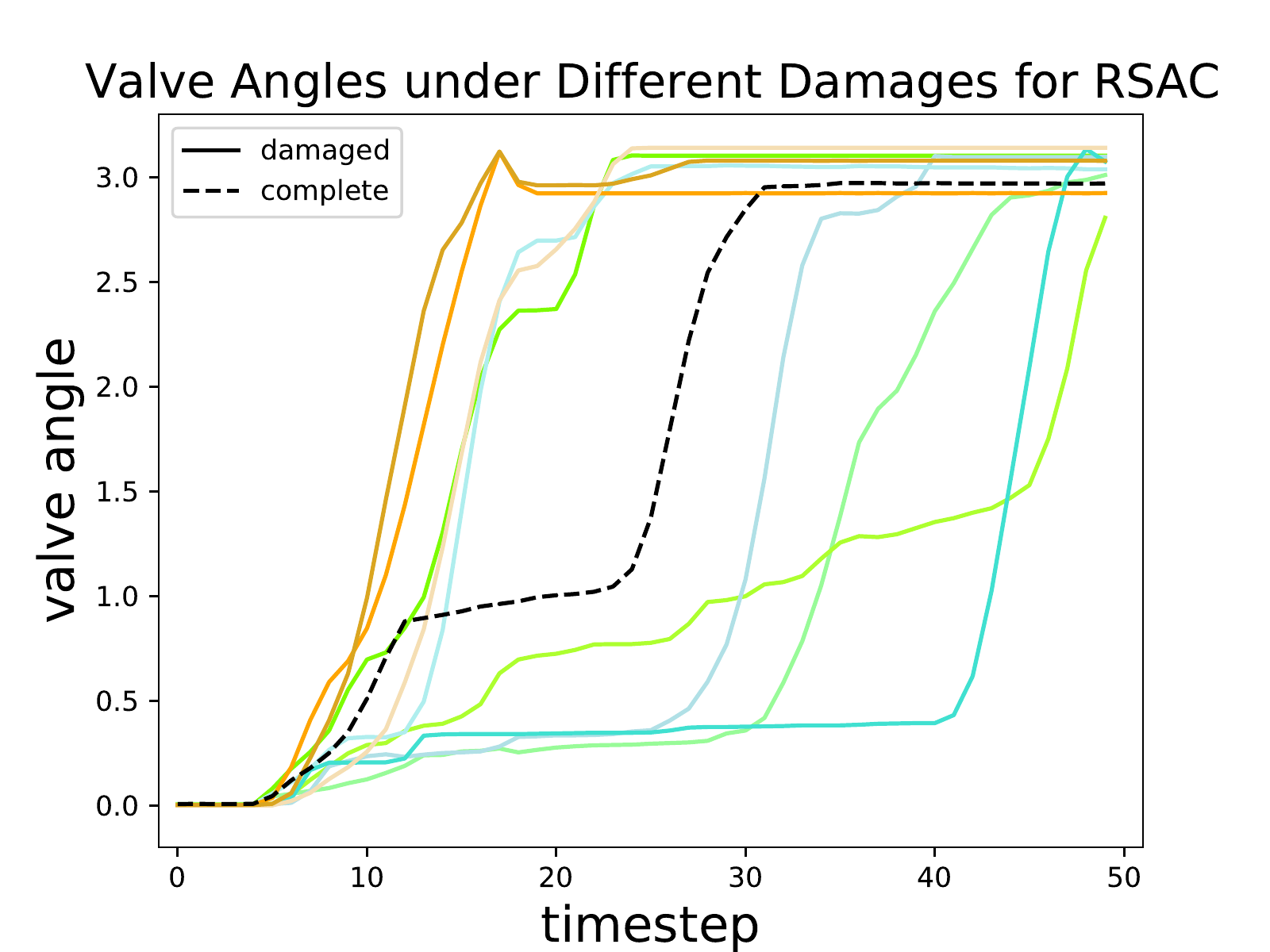}
        \label{fig:at_nine}
    }
    \subfigure[]{
	\includegraphics[width=4cm]{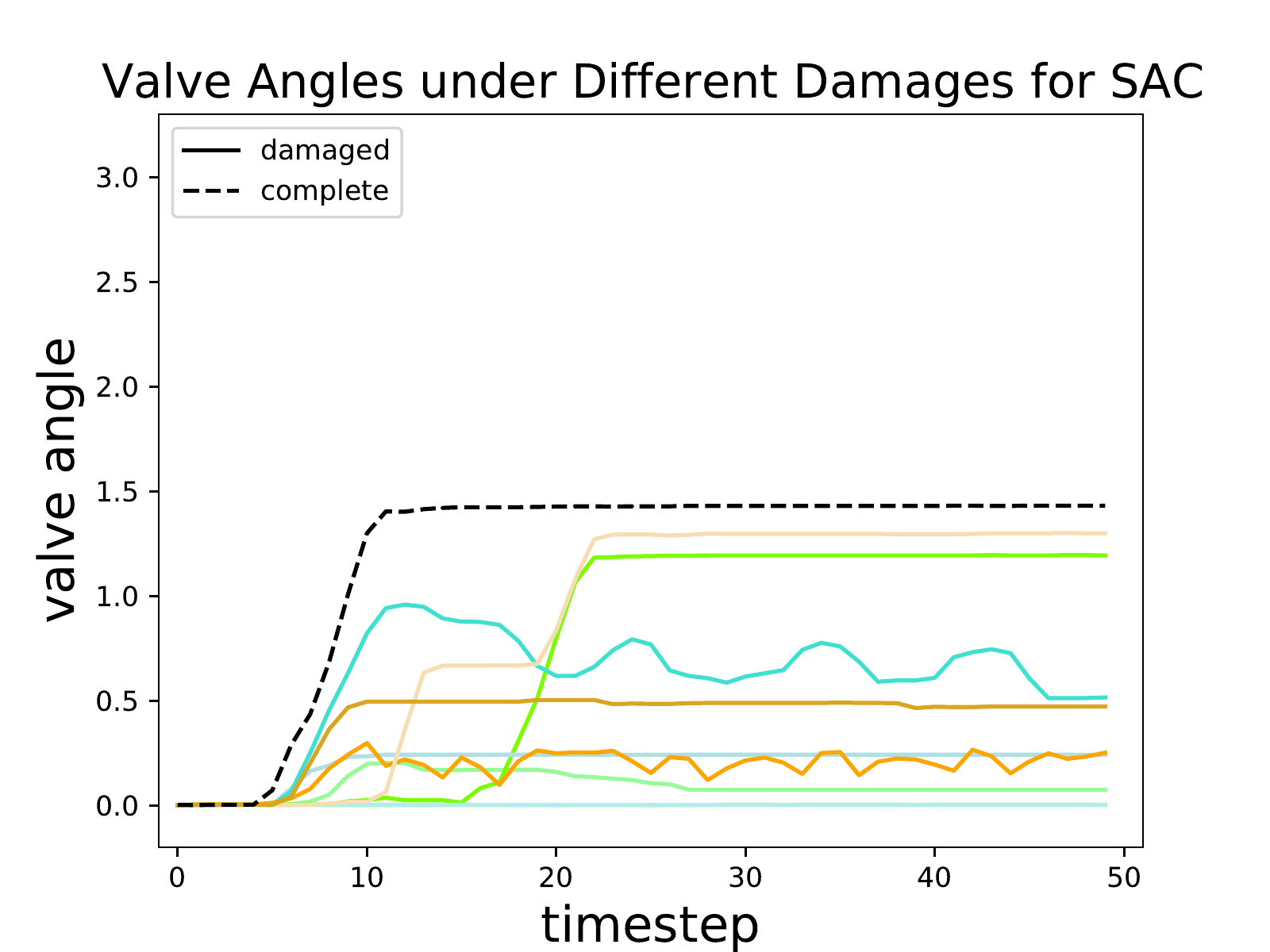}
        \label{fig:pure_baseline}
    }
    \caption{The figure shows the change of valve angles under different joint damage cases. Solid lines indicate different damage cases, while black dashed lines indicate complete cases.}
    \label{fig:nine_cases}
\end{figure}
In this section, we will first compare the performance of our algorithm with a standard reinforcement learning algorithm under joint damage cases. Then we will also demonstrate that our algorithm can also yield increased robustness other than joint damage, for example, noise resistance.
\subsection{Experiment Platform}
We evaluate our algorithm on ROBEL platform\cite{ahn2020robel}. ROBEL consists of two robot platforms: (1) D'Claw as the manipulation platform and (2) D'Kitty as the locomotion platform.

In our manipulation experiment, the D'Claw robot is trying to turn a valve from $0^\circ$ to $180^\circ$. The task has dense rewards based on the difference between the current valve angle and the target angle. The task is considered successful when the valve is turned to more than $170^\circ$.
% The robot is shown in Figure \ref{fig:dclaw}:.
% \begin{figure}
%     \centering
%     \includegraphics[width=6cm]{pic/dclaw.jpeg}
%     \caption{D'Claw robot used in the experiment}
%     \label{fig:dclaw}
% \end{figure}

In our locomotion experiment, the D'Kitty robot is trying to move from its current Cartesian position to the desired position. It also has dense rewards based on the distance to the goal. The task is considered successful when it is within $0.5$ meters of the destination.
% More specifically, its reward function is defined as follows:
% \begin{equation}
%     r_t = -5|\delta \alpha_{t,obj}| - ||\mathbf{q}_{nominal} - \mathbf{q}|| - ||\mathbf{\dot{\mathbf{q}}_t}|| + 10\mathbbm{1}(|\delta \alpha_{t,obj}| < 0.25) + 50\mathbbm{1}(|\delta\alpha_{t,obj}| < 0.1)
% \end{equation}
\subsection{Experiment Setup}
\begin{figure*}[htbp]
    \centering
    \includegraphics[width=13cm]{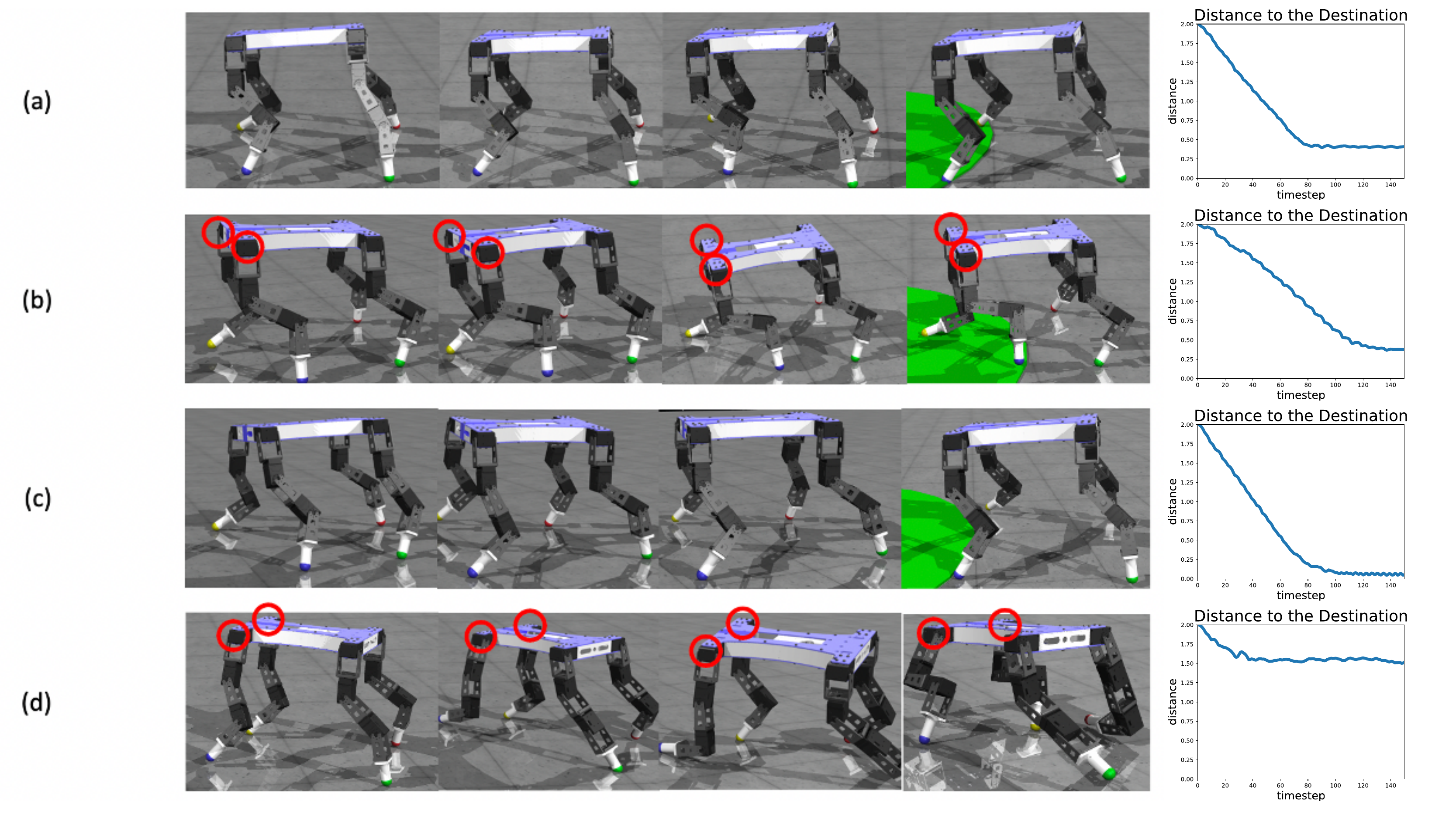}
    \caption{Sample figure of D'Kitty locomotion experiements of single-joint-damaged cases and complete cases. Photos are not sampled with the same time interval, but for clarity. Figure (a) shows the RSAC in a complete case. Figure (b) shows RSAC in a damage case. Figure (c) shows the SAC in a complete case. Figure (d) shows the SAC in a damage case.}
    \label{fig:kitty_exp}
\end{figure*}
In our experiments, if a joint is damaged, it is set to a random joint angle and any control signal will not be able to further change it. Angles which would make the tasks impossible to complete are removed, \eg angles that will inevitably make the valve get stuck. Moreover, joint angle sensors for damaged ones are not available for the robot and would always return 0 by default. 

Note that our algorithm can also have increased robustness if joint malfunction is defined as a joint executing random action. % ToDO: show in appendix

Soft Actor-Critic(SAC)\cite{haarnoja2018soft} is chosen as the reinforcement learning algorithm for the agent. 
In our cases, because the agent needs to explore a wide range of action space to gather experience for a better solution under damage cases. If a traditional RL algorithm is implemented, \eg, DDPG\cite{lillicrap2015continuous}, the agent is likely to over explore a certain damage case. When the damage case changes, it will not be able to explore spaces where the best solutions exist. On the contrary, in SAC, an entropy term is implemented to encourage the exploration process, which relieves the problem above. Our proposed algorithm is referred to as Robust Soft Actor-Critic(RSAC) in the following sections.
\subsection{D'Claw experiments}
Because of the configuration of the D'Claw robot, there exist cases when three joints are damaged simultaneously that would make the task impossible to complete. Therefore, D'Claw manipulation experiments are designed to demonstrate the robustness when arbitrary one or two joints are damaged. 
\subsubsection{Single-Joint-Damaged Experiment}
Single-joint damaged experiments are designed to show details of RSAC methods compared to traditional SAC.
The experiments of single-joint-damaged cases are shown in Figure \ref{fig:nine_cases}. Detailed photos of the experiment processes are shown in Figure \ref{fig:big_exp}. Both RSAC and SAC algorithms are trained in simulation and directly applied to an actual robot without any fine-tuning. There are differences in the valve position and dynamics for the real environment, while the robot does not have any vision sensor to infer about valve position, which greatly threatens the success rate of the algorithm without fine-tuning. SAC cannot finish the task even when its joints are complete. While RSAC has great performances in each single-joint-damaged case, which demonstrates evidently increased robustness to joint damage compared to SAC. Moreover, in Figure \ref{fig:at_nine}, valve angles change significantly differently under different damage cases, which implies control policies are able to adapt to corresponding scenarios to some extent rather than simply apply the original one.
\begin{figure}[htbp]
    \centering
    \subfigure[Success rate for RSAC in simulation.]{
        \includegraphics[width=4cm]{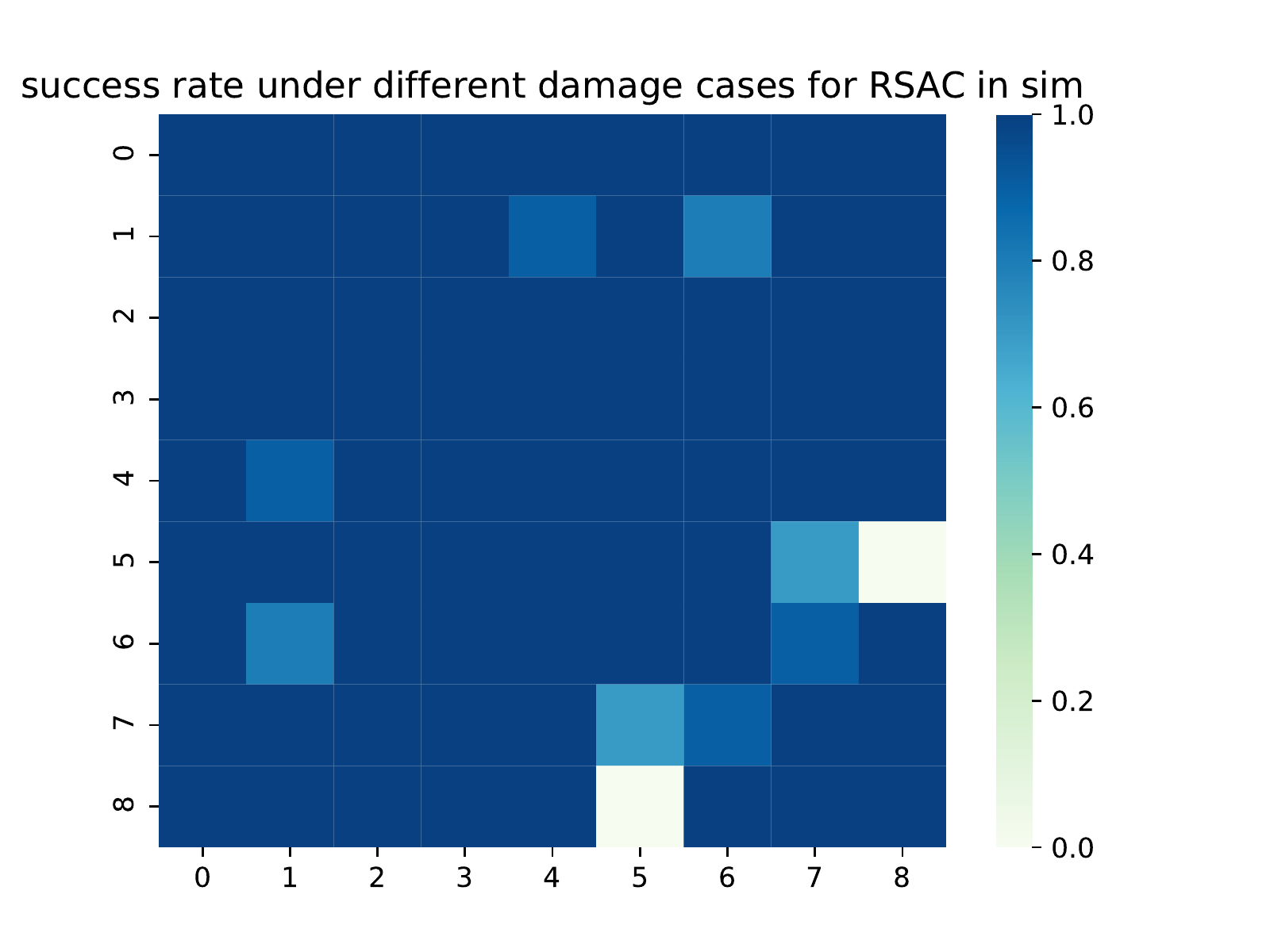}
        \label{fig:rsac_heatmap_sim}

    }
    \subfigure[Success rate for SAC in simulation.]{
	\includegraphics[width=4cm]{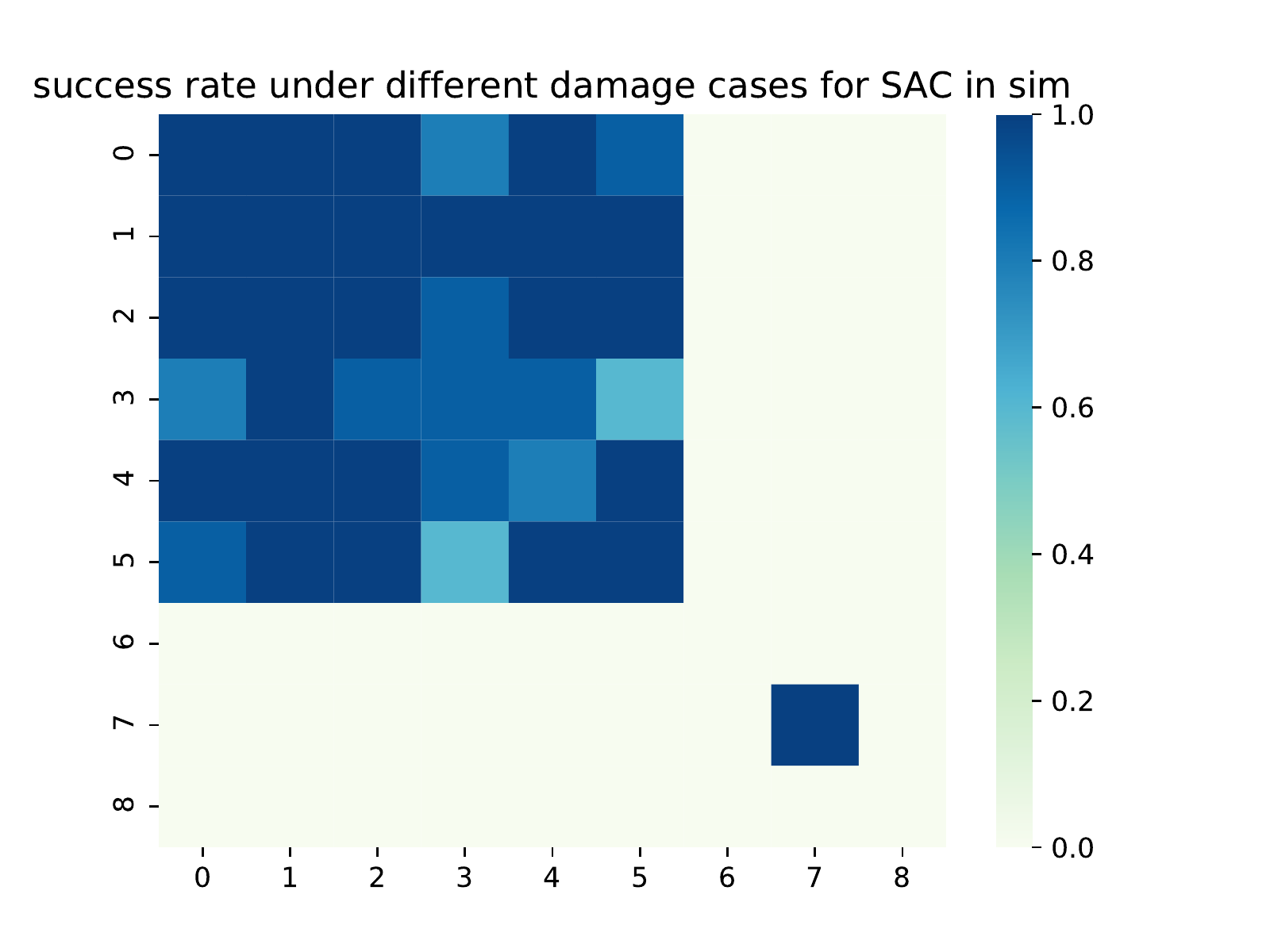}
        \label{fig:sac_heatmap_sim}
    }
    \quad
    \subfigure[Success rate for RSAC on a real robot.]{
        \includegraphics[width=4cm]{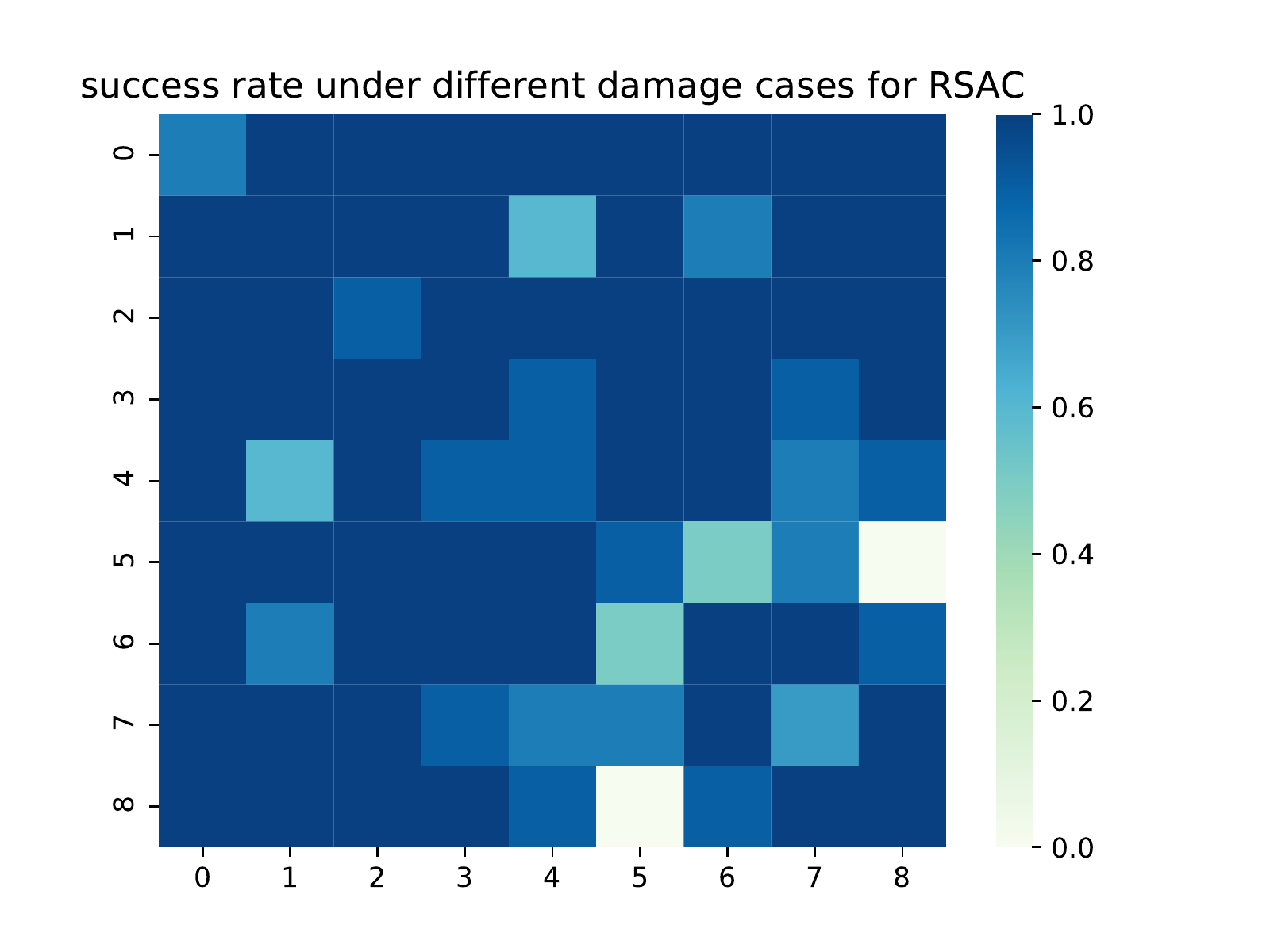}
        \label{fig:rsac_heatmap_real}
    }
    \subfigure[Success rate for SAC on a real robot.]{
	\includegraphics[width=4cm]{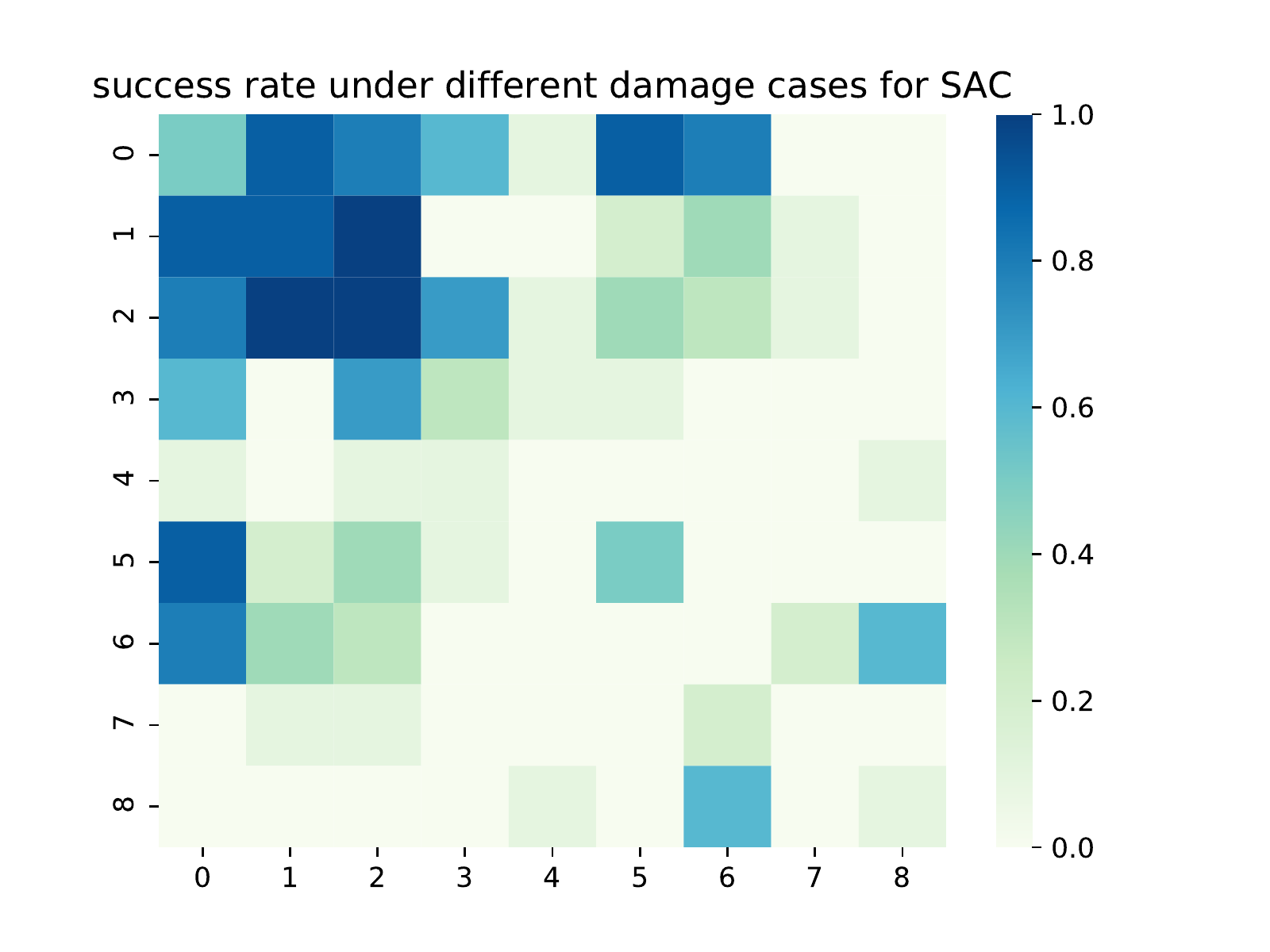}
        \label{fig:sac_heatmap_real}
    }
    \caption{The figure shows the success rate under different joint damage cases. The element in the ith row and jth column indicates the success rate when joint i and j are damaged. Each damage case is evaluated ten times on the D'Claw robot to get the success rate.}
    \label{fig:heatmap}
\end{figure}

\subsubsection{Multiple-Joint-Damaged Experiment}
Based on the previous experiment, a more complete experiments is conducted to compare two algorithms thoroughly\footnote{We used different damping parameters for the valve compared to the previous experiment}. Experiment results of success rate for arbitrary combinations of two damaged joints are shown in Figure \ref{fig:heatmap}. Experiments were conducted in both simulation and the real world. For each case, ten experiments were conducted to evaluate the success rate. According to the experiments, RSAC is capable of adapting to most damage cases, while SAC fails on the contrary. More specifically, the learned policy in SAC seems to rely heavily on joint 3 to joint 8, while damage to joint 0 to 2 has limited influence. But RSAC does not have such relying situations, which results in the increased robustness.

Moreover, the success rate between simulation experiments and real-world experiments for RSAC is pretty similar, while SAC doesn't have such properties. It shows that RSAC has an ability to handle sim-to-real problems. It has a significant meaning to our RSAC algorithm. Since robot malfunction scenarios usually lead to unpredictable results,  which is very dangerous and is likely to further damage the robot. It would be inappropriate if we train RSAC using a real robot. The minor sim-to-real gap enables us to train RSAC in pure simulation, the algorithm can be deployed to real robots without or with little fine-tuning.

Moreover, the increase between RSAC simulation results and SAC simulation results indicate that the increase in the real world not only arises from robustness over the sim2real gap but also results from its adaptation to different damage cases.
\vspace{-0.03cm}
\subsubsection{Noise Resistance Experiment}
We also demonstrate that not only joint damage cases, our method also increases robustness in other aspects. In this experiment, Gaussian noise is added to the action space.
More concretely, the action $\mathbf{a}_t'$ our robot conducts is defined as $\mathbf{a}_t' = \mathbf{a}_t + \delta$, where $\delta \sim \mathcal{N}(0,1)$. Both RSAC and SAC were evaluated 30 episodes on a real robot to get success rates. RSAC succeeded in 76.67\% cases, while SAC had only a 10.00\% success rate.

\subsection{D'Kitty Experiment}
\vspace{-0.1cm}
In the D'Kitty locomotion experiment, experiments are designed to demonstrate the robustness when arbitrary one or two joints are damaged. Our experiments are conducted in simulation. D'Kitty is more challenging than locomotion tasks mentioned in previous robot damage work. It has fewer legs, and its center of mass is higher, which makes it more unstable. D'Kitty locomotion experiments are much more difficult than D'Claw manipulation experiments. There are a few reasons for it: (1) A wrong choice in locomotion would lead to the robot falling, thus following actions are not possible to achieve; (2) If a joint is damaged in our experiment, its joint angle information is no longer available for the RL agent. In manipulation problems, the agent could choose not to use the finger damaged, while in locomotion problems, the robot will still have to rely on those damaged legs to keep balance. The unavailability of joint angle information would make it much more challenging. 

\begin{figure}[htbp]
    \centering
    \subfigure[Success rate for RSAC in simulation.]{
        \includegraphics[width=4cm]{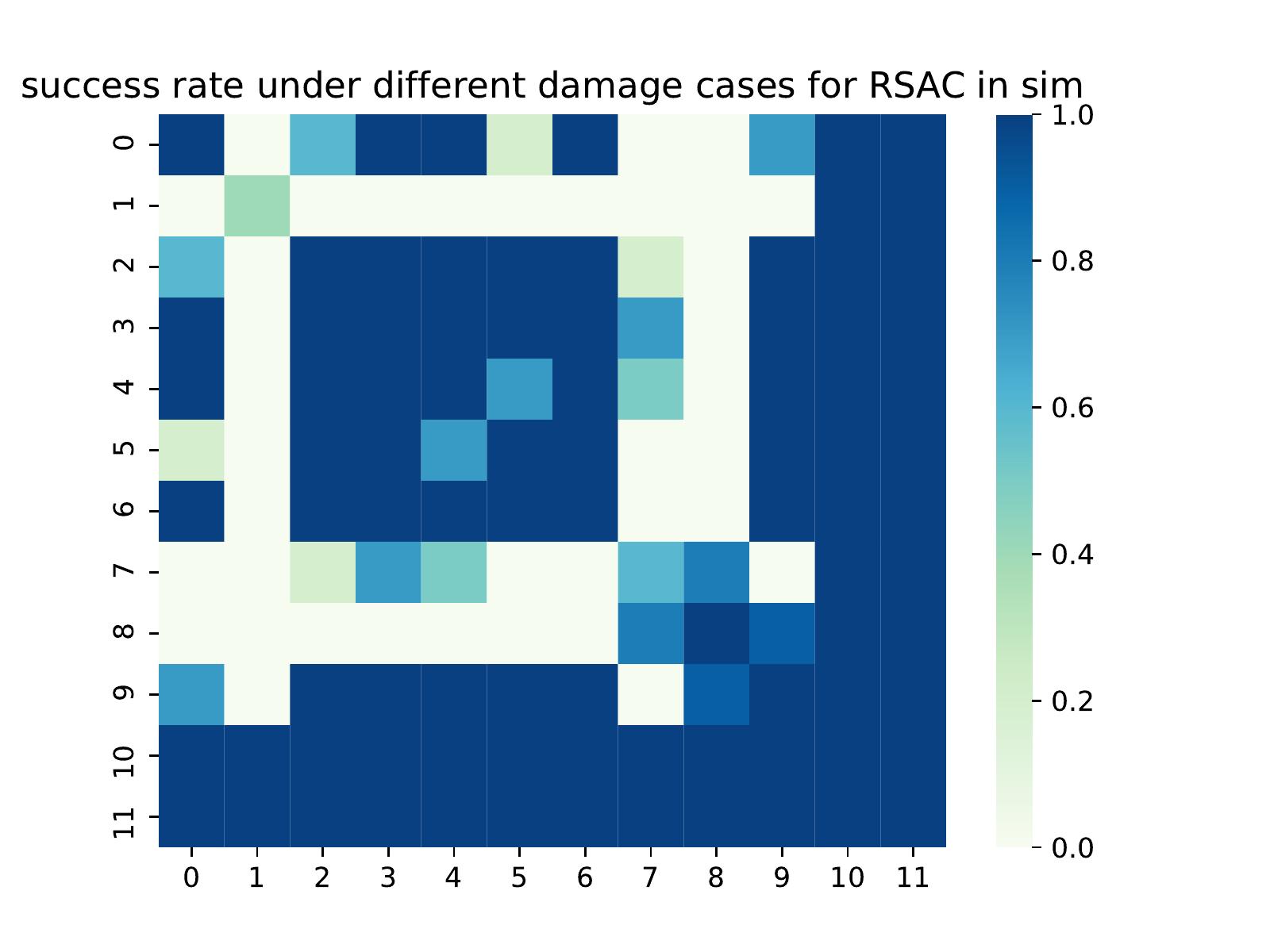}
        \label{fig:rsac_heatmap_kitty}

    }
    \subfigure[Success rate for SAC in simulation.]{
	\includegraphics[width=4cm]{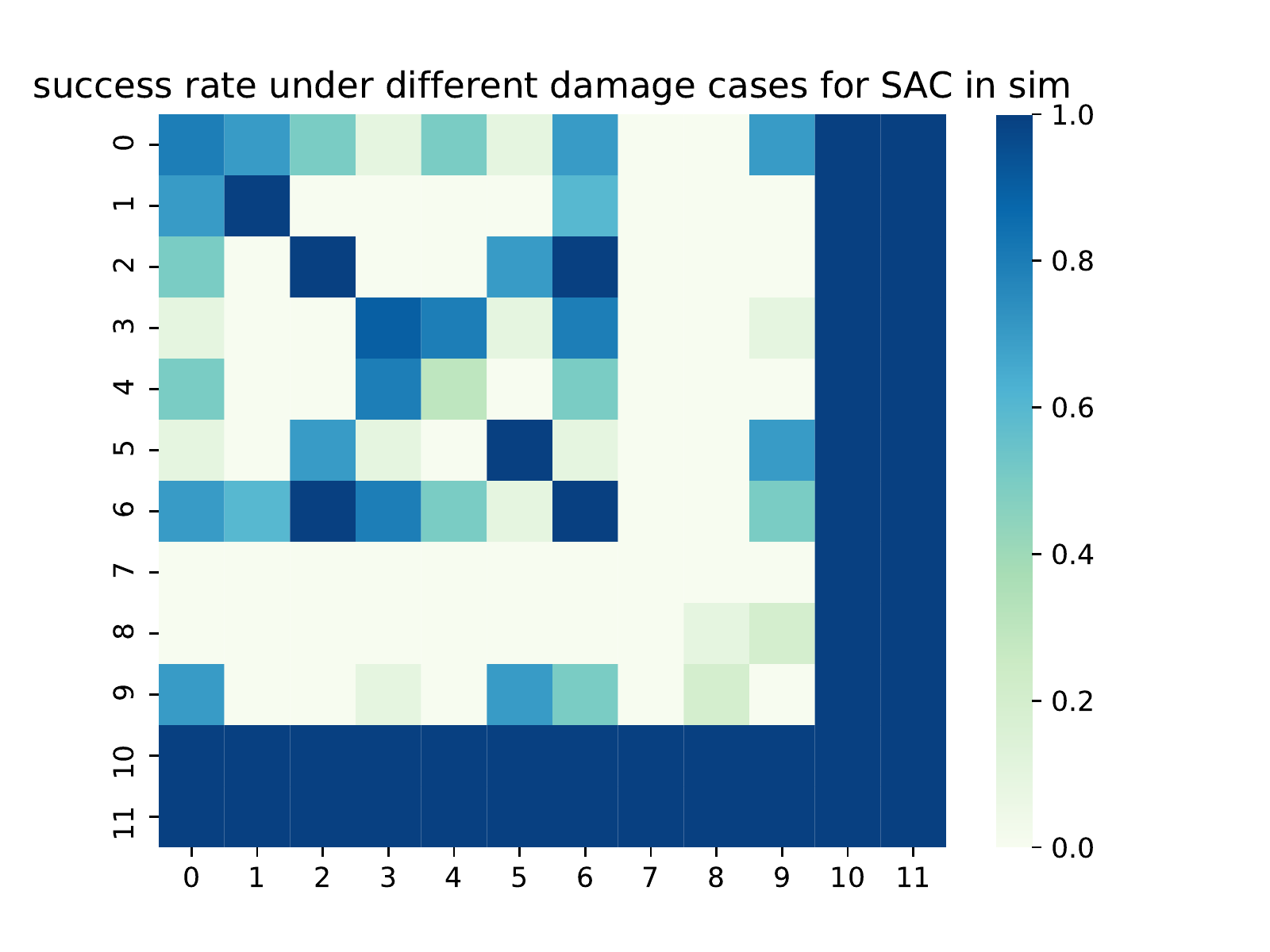}
        \label{fig:sac_heatmap_kitty}
    }
    \caption{The figure shows the success rate under different joint damage cases. Both experiments are conducted in simulation environments. The element in the ith row and jth column indicates the success rate when joint i and j are damaged. Each damage case is evaluated ten times on the D'Kitty robot to get the success rate.}
    \label{fig:kitty_heatmap}
\end{figure}

However difficult the task is, our RSAC method still has an increased success rate over baseline. The success rate when arbitrary two joints are damaged is shown in \ref{fig:kitty_heatmap}.
\vspace{-0.5cm}
\section{Conclusion}
% \vspace{-0.5cm}
Solving robot damage problems, especially joint malfunction, is of great importance in the robotics community. We proposed an adversarial reinforcement learning framework, in which the RL agent is trained iteratively under the most challenging robot damage situations. The robot is fault-aware for its awareness of joint working states. We implemented our algorithm in manipulation and locomotion scenarios. Experiments demonstrated a substantially increased robustness, not only over joint damages but also noise resistance and simulation-real differences.
% \small{
% \section*{ACKNOWLEDGMENT}

% }
\bibliographystyle{IEEEtran} % (0.5 page, about 20 refs)
\bibliography{IEEEabrv,ref}

\end{document}